\documentclass[conference]{IEEEtran}
\IEEEoverridecommandlockouts

\usepackage{cite}
\usepackage{amsmath,amssymb,amsfonts}
\usepackage{graphicx}
\usepackage{textcomp}
\usepackage{xcolor}
\usepackage{url}
\usepackage{booktabs}
\usepackage{multirow}
\usepackage{dblfloatfix}
\usepackage{algorithm}
\usepackage{algpseudocode}
\usepackage{subcaption}
\def\BibTeX{{\rm B\kern-.05em{\sc i\kern-.025em b}\kern-.08em
    T\kern-.1667em\lower.7ex\hbox{E}\kern-.125emX}}
\begin{document}

\title{Yet Another Concept Drift Benchmark? A Principled Framework for Evaluating Drift Detectors}

\author{\IEEEauthorblockN{Lorenzo Iovine}
\IEEEauthorblockA{\textit{DEIB},
\textit{Politecnico di Milano}\\
Milan, Italy \\
lorenzo.iovine@polimi.it}
\and
\IEEEauthorblockN{Giacomo Ziffer}
\IEEEauthorblockA{\textit{DEIB},
\textit{Politecnico di Milano}\\
Milan, Italy \\
giacomo.ziffer@polimi.it}
\and
\IEEEauthorblockN{Emanuele {Della Valle}}
\IEEEauthorblockA{\textit{DEIB},
\textit{Politecnico di Milano}\\
Milan, Italy \\
emanuele.dellavalle@polimi.it}
}

\maketitle

\begin{abstract}
Concept drift detection is a fundamental component of data stream learning, yet its empirical evaluation remains based on loosely defined protocols, ambiguous alarm semantics, and inconsistent metrics. We propose a principled benchmark grounded in an explicit formulation of drift observability and alarm semantics. Our framework introduces finite detection windows and repeated alarms, enabling a clear separation between detection accuracy, timeliness, and stability, and employs a reproducible observer to provide a controlled and interpretable performance signal. Building on this formulation, we define normalized evaluation metrics and a composite score addressing limitations of existing measures. We validate the framework through a large-scale stress-testing evaluation of 10 detectors across 9 learner configurations and 16 datasets.
\end{abstract}


\section{Introduction}
\label{section: intro}
In data stream learning, \textit{concept drift} is commonly referred to as situations in which the predictive relationship between inputs and output changes over time. A widely adopted view characterizes drift as a change in the underlying decision boundary, typically formalized as a change in the conditional distribution $p(y|x)$~\cite{WidmerK1996,GamaZBPB2014}. Such changes take different forms, including abrupt, gradual, incremental, or recurring~\cite{Tsymbal2004, WebbHCNP2016}.

\begin{figure*}[t]
\centering
    \includegraphics[width=1.0\linewidth]{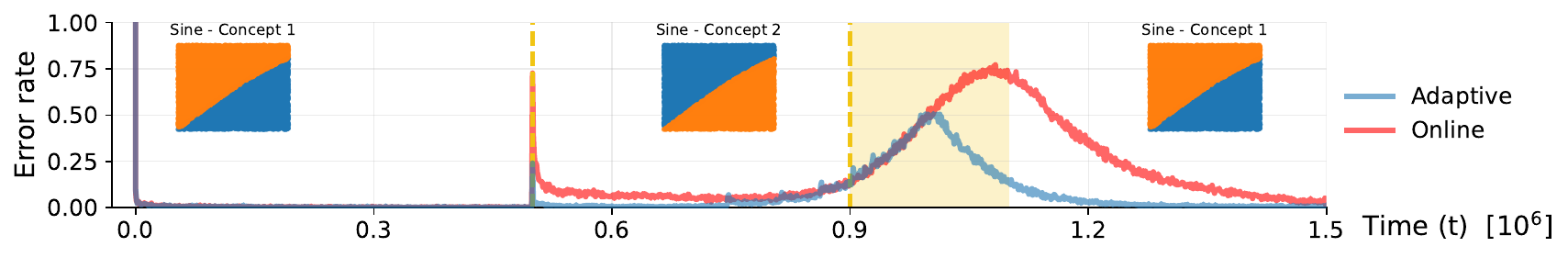}
    \caption{
        Emblematic case of comparison of error rates over time in non-stationary conditions between a HoeffdingTree~\cite{HultenSD2001} (red) and a HoeffdingAdaptiveTree~\cite{BifetG2009} (blue) on a stream generated with SineGenerator.
    }
    \label{fig:performance}
\end{figure*}

We adopt an operational perspective grounded in performance monitoring. Since the decision boundary is not directly observable in real-world scenarios~\cite{HoensPC2012, KhamassiSHG2018}, drift must be inferred from observable quantities. Figure~\ref{fig:performance} illustrates this indirect observability through the error rates of predictive models over time. Consequently, concept drift detection is inherently a performance-based inference problem: detectors do not observe drift itself, but rely on statistics derived from a learner, which acts as observational instrument.

\textbf{The problem} we face is that the empirical evaluation of drift detectors remains poorly standardized and conceptually fragile~\cite{LuLDGGZ2018}. Existing benchmarks~\cite{MahgoubMEA2022} rely on heterogeneous practices, unclear alarm semantics, and performance measures that are difficult to compare across studies. This raises a fundamental question: \emph{how reliable are current evaluation practices for concept drift detection?}. We argue that answering this question requires first separating detector behavior from the confounding effects introduced by the predictive learner. Our position is not that learner dependence should be ignored, but that it should first be studied under a controlled observational setting. Only once detector behavior is characterized independently of uncontrolled learner dynamics can the learner itself be reintroduced as an explicit factor of analysis.

Our work is driven by the following \textbf{research question}: \emph{how can concept drift detectors be evaluated and compared in a principled and reproducible way?}
Our \textbf{contributions} are: (i) a \textbf{systematic analysis} of current benchmarking practices, identifying key failure modes and their impact on empirical conclusions; (ii) a principled \textbf{evaluation framework} for concept drift detection grounded in an explicit notion of observability; (iii) a principled \textbf{alarm semantics} that enables a consistent interpretation of detection events over time; (iv) a set of normalized \textbf{performance metrics} that disentangle detection accuracy, timeliness, and stability; (v) \textbf{Sliding Heatmap}, a reproducible observer that computes class counts over a fixed-size window of recent observations; (vi) an \textbf{extensive empirical evaluation} on synthetic and real-world data streams. To support full reproducibility of our benchmark, we openly release the complete code base of our evaluation framework as open-source executable artifacts.~\footnote{The code to reproduce all experiments is publicly available at: \url{https://github.com/lorenzoiovine99/Benchmarking_Framework_for_Detectors}}

The remainder of the paper is organized as follows. Section~\ref{section: fragility} analyzes limitations of current evaluation practices. Section~\ref{section: formulation} formalizes the problem. Section~\ref{section: metrics} introduces the proposed metrics. Section~\ref{section: learner} presents Sliding Heatmap. Section~\ref{section: results} reports the experimental results. Finally, Section~\ref{section: conclusions} concludes the paper.

\begin{table}[b]
\caption{Drift detector performance on SINE Gradual under three
observers. Bold indicates the highest MTR for each observer.}
\label{tab:sota_comparison}
\centering
\footnotesize
\setlength{\tabcolsep}{6pt}
\renewcommand{\arraystretch}{1.08}
\begin{tabular}{@{}llrrrr@{}}
\toprule
\textbf{Observer} & \textbf{Detector} & \textbf{MTFA} &
\textbf{MTD} & \textbf{MDR} & \textbf{MTR} \\
\midrule
\multirow{3}{*}{NB}
& ADWIN & 9728   & 518.4  & 0.4 & 10.4 \\
& CUSUM & 91195  & 2857.5 & 0.6 & 14.2 \\
& DDM   & 238456 & 2688.7 & 0.7 & \textbf{29.6} \\
\midrule
\multirow{3}{*}{HT}
& ADWIN & 6846   & 603.4  & 0.2 & 8.8 \\
& CUSUM & 70801  & 974.8  & 0.4 & \textbf{40.4} \\
& DDM   & 1408.3 & 711.2  & 0.4 & 1.1 \\
\midrule
\multirow{3}{*}{HAT}
& ADWIN & 4365   & 405.3  & 0.3 & 7.2 \\
& CUSUM & 28891  & 840.5  & 0.3 & \textbf{22.9} \\
& DDM   & 23130  & 358.0  & 0.8 & 14.4 \\
\bottomrule
\end{tabular}
\end{table}

\section{On the Fragility of Concept Drift Evaluation}
\label{section: fragility}
Despite the widespread use of concept drift detectors, their empirical evaluation remains inconsistent across studies. Existing benchmarks differ substantially in their choice of base learner, evaluation protocol, and performance metrics.  While often treated as implementation details, these choices implicitly determine how drift is observed, how detector outputs are interpreted, and how performance is quantified. To assess their impact, we reproduce representative evaluation protocols from prior work, including classical studies~\cite{Goncalves2014}, synthetic error-based evaluations~\cite{Bifet2017}, standardized frameworks~\cite{HaugTK2022}, and recent large-scale benchmarks~\cite{MahgoubMEA2022, MahdiBS2024, TosiT2024}. Our analysis reveals fundamental failure modes that undermine the reliability, comparability, and interpretability of existing evaluations.

\subsection{Learner-Induced Bias and Metric Instability}
A critical but often overlooked factor is the base learner used to generate the performance signal. To quantify this effect, we evaluate a set of drift detectors using as learners: Naive Bayes (NB), Hoeffding Tree (HT), and Hoeffding Adaptive Tree (HAT), on SINE Gradual dataset, using standard metrics.

We consider Mean Time to Detect (MTD), measuring detection delay expressed in number of instances; Mean Time between False Alarms (MTFA), measuring the average number of instances between consecutive false alarms; Missed Detection Rate (MDR), the proportion of undetected drifts; and the Mean Time Ratio (MTR), defined as:
\[
\mathrm{MTR} = \frac{\mathrm{MTFA}}{\mathrm{MTD}} \cdot (1 - \mathrm{MDR}).
\]

Table~\ref{tab:sota_comparison} illustrates how observer choice affects the comparison. In addition, MTR depends on the mean interval between false alarms and is undefined when fewer than two false alarms occur. These limitations motivate specifying the observer and the scoring rules explicitly.

\subsection{Ambiguity in Detection Semantics and Evaluation Protocols}
The issues discussed above are amplified by ambiguities in the definition and interpretation of detection events. Existing benchmarks lack a consistent and explicit notion of detection semantics, including how alarms are associated with drift events and how multiple alarms are handled. A central issue concerns repeated alarms. In most prior work, alarms triggered after an initial detection are either ignored or counted as independent false positives. However, without a principled definition of drift boundaries, their interpretation remains unclear. In practice, repeated alarms often correspond to the same underlying degradation period, yet current protocols do not distinguish them from spurious detections in stationary phases. This ambiguity is closely related to the definition of the detection window. In existing benchmarks, this window is typically fixed a priori or implicitly determined by the experimental setup. Because the window determines which alarms can be credited to a drift, its length must be specified as part of the evaluation protocol.

\section{Problem Formulation}
\label{section: formulation}
Concept drift detection has been formalized in different ways in the literature. However, as discussed in Section~\ref{section: fragility}, existing evaluation practices rely on implicit and often inconsistent assumptions about observability, detection semantics, and performance measurement. A common and theoretically grounded formulation is based on the Bayesian decomposition of the posterior distribution~\cite{Zliobaite2010}.

\subsection{Concept Drift and Observability}
\begin{figure*}[t]
    \centering
    \begin{subfigure}{0.45\textwidth}
        \centering
        \includegraphics[width=\linewidth]{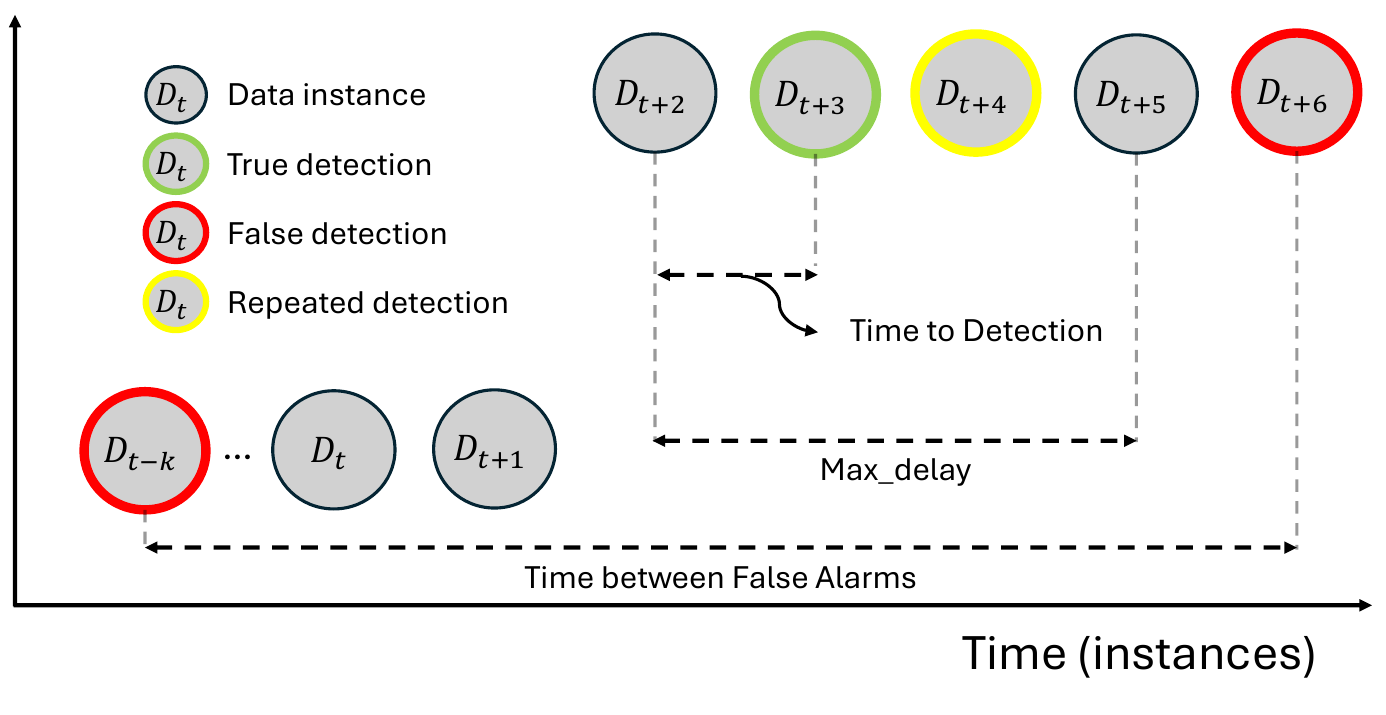}
        \caption{Abrupt Concept Drift}
        \label{fig:abrupt}
    \end{subfigure}\hfill
    \begin{subfigure}{0.45\textwidth}
        \centering
        \includegraphics[width=\linewidth]{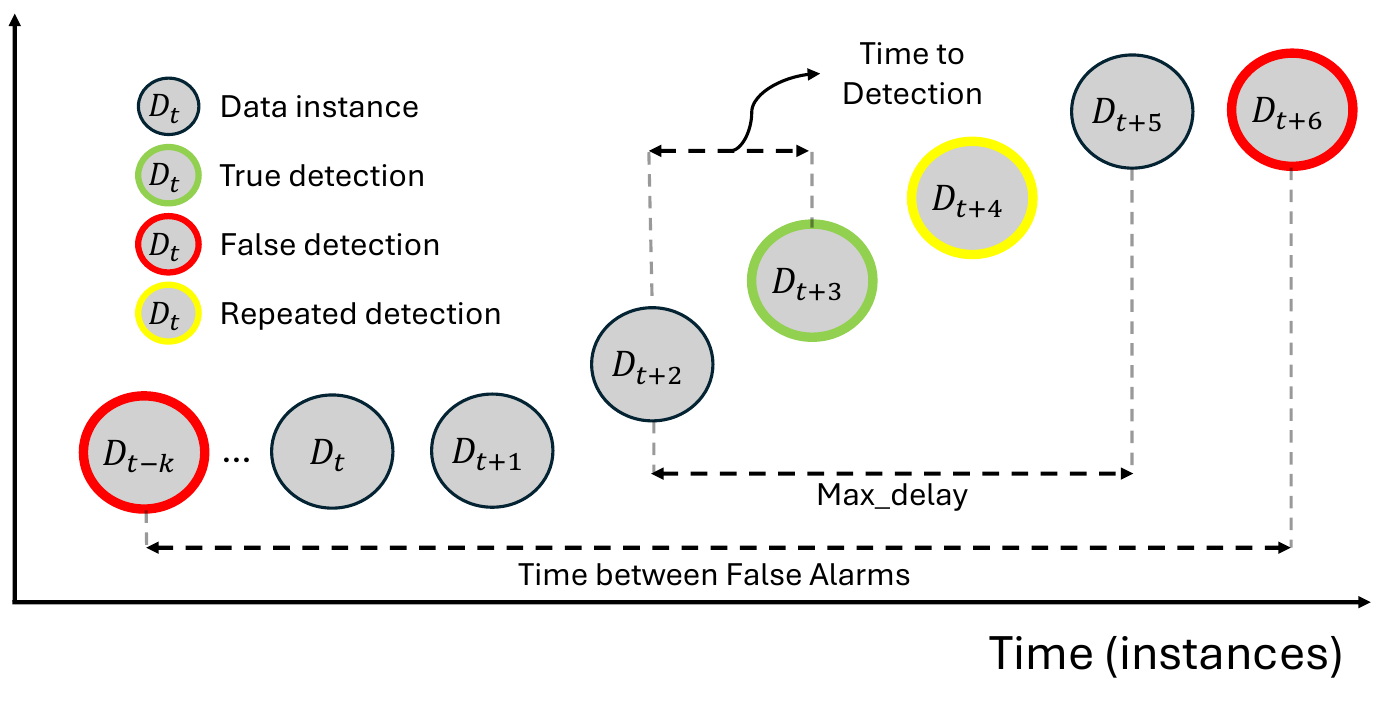}
        \caption{Gradual Concept Drift}
        \label{fig:gradual}
    \end{subfigure}
    \caption{
        Visualization of detector alarms under an abrupt (a) and a gradual (b) concept drift. Green circles denote true detections, yellow circles repeated alarms, and red circles false alarms. The figures highlight time to detection, maximum allowed delay (max\_delay), and time between false alarms.
    }
    \label{fig:visualization}
\end{figure*}

In a supervised classification setting, predictions are governed by the posterior distribution $p(y|x)$, which can be expressed using Bayes' theorem as:
\begin{equation}
    p(y|x) = \frac{p(x|y) p(y)}{p(x)}.
\end{equation}

\textbf{Real concept drift} is a change in \(p(y\mid x)\). \textbf{Virtual drift} is a change in \(p(x)\) while \(p(y\mid x)\) remains unchanged. A change in the class prior \(p(y)\) or the class-conditional distribution \(p(x\mid y)\) can also change \(p(y\mid x)\).

Crucially, $p(y|x)$ and the associated decision boundary are \textbf{not directly observable} in real-world streams (Section~\ref{section: fragility}). Drift must therefore be inferred indirectly through observable quantities, typically the predictive performance of a classifier. In this setting, the learner acts as an observational instrument, making changes in $p(y|x)$ measurable through performance variations. This dependency makes drift detection learner-dependent and a key source of variability in empirical evaluation. 
From an operational perspective, only \textbf{performance degradations} are actionable, while improvements should not trigger alarms. Drift detection is therefore a \textbf{one-sided problem}, where detectors respond only to statistically significant increases in error.

\subsection{Alarm Semantics and Detection Window}
The ambiguity of current evaluation protocols is largely due to the lack of a formal definition of how detector output should be interpreted over time. Since drift is inferred through performance degradation, detectors raise alarms when a monitored statistic exhibits a statistically significant increase. Interpreting these alarms is non-trivial, especially when multiple alarms occur around the same drift event or in stationary phases. We formalize alarm semantics by explicitly relating detector outputs to known drift events in a time-based framework. Given a stream with annotated drift points, each alarm is categorized as follows: \textbf{true detection:} an alarm within a predefined window after drift onset; \textbf{false alarm:} an alarm outside any valid window; \textbf{repeated alarm:} additional alarms occurring after a true detection and within the same window. Distinguishing repeated alarms is essential: without it, detectors that trigger multiple times for the same drift would be unfairly penalized.

To associate alarms with drift events, we define a \textbf{detection window} that starts at the annotated drift time and extends to a maximum number of instances, denoted \textbf{max\_delay}. The max\_delay parameter defines the finite horizon over which alarms are associated with a drift event. In our experiments, we set the max\_delay to the Sliding Heatmap buffer size \(W\). When windows overlap, alarms are associated with the earliest drift. Although these semantics apply uniformly, different types of drift induce different patterns (Figures~\ref{fig:abrupt} and~\ref{fig:gradual}).

\section{Metrics for Concept Drift Detection}
\label{section: metrics}
Evaluating concept drift detectors is challenging, as detectors trade off detection delay, false alarms, and missed detections. Commonly used metrics suffer from lack of boundedness, instability under few false alarms, and poor comparability across datasets. 
We present a set of performance metrics that align with the defined alarm semantics and are designed to be well-defined, bounded, and comparable across datasets.

\subsection{Detection Performance Metrics}
We define metrics that explicitly account for true detections, false alarms, and repeated alarms. Repeated alarms are identified but not treated as false alarms; they are not aggregated into a standalone metric. This avoids a key limitation of existing benchmarks, where repeated alarms are conflated with false positives, leading to misleading evaluations. All metrics are bounded in $[0,1]$, enabling direct comparison across detectors and datasets. We refer to \emph{effective alarms} as non-repeated alarms, i.e. true detections or false alarms.
Although repeated alarms rely on drift localization, this is not a limitation in our setting: in controlled benchmarks, drift points are known, enabling unambiguous evaluation. In practice, repeated alarms can be approximated via post-drift stabilization or human validation. This distinction also reflects common practice, where multiple alarms for the same drift are typically ignored rather than treated as independent false positives.

\paragraph{True Detection Rate (TDR)}
\begin{equation}
    TDR = \frac{\#\text{ true detections}}{\#\text{ drifts}}.
\end{equation}

\paragraph{False Alarm Rate (FAR)} We normalize FAR by effective alarms to isolate spurious alarms from repeated detections.
\begin{equation}
    FAR = \frac{\#\text{ false alarms}}{\#\text{ effective alarms}}.
\end{equation}

\paragraph{Mean Time to Detect - Normalized ($\mathrm{MTD}_{\mathrm{norm}}$)}
\begin{equation}
    \mathrm{MTD}_{\mathrm{norm}} = \frac{\mathrm{MTD}}{\mathrm{max\_delay}}.
\end{equation}

\subsection{Composite Metric for Overall Evaluation}
While individual metrics provide detailed insights, practical benchmarking often requires a single score summarizing overall performance. Such a metric should balance detection accuracy, false alarms, and delay, while remaining interpretable and bounded. We summarize detector performance with the bounded composite score:
\begin{equation}
    M_1={TDR\cdot(1-FAR)\cdot(1-\mathrm{MTD}_{\mathrm{norm}})}.
\end{equation}
When no annotated drift is detected, we set $M_1=0$ directly, since $MTD_{norm}$ is undefined. We report the component metrics separately to show the underlying trade-offs. While not explicitly quantified here, repeated alarms can be incorporated into additional stability measures or extended composite metrics.

For each detector and stream, we compute the coefficient of variation of \(M_1\) across the tested Sliding Heatmap configurations:
\begin{equation}
    CV(M_1)=\frac{\sigma(M_1)}{\mu(M_1)},
\end{equation}
where $\sigma(M_1)$ and $\mu(M_1)$ denote the standard deviation and mean across configurations.

\section{A Windowed Observer for Drift Benchmarking}
\label{section: learner}
Sliding Heatmap extends the Histogram Rule [18] to streams by maintaining class counts in a grid over a FIFO buffer of \(W\) observations. It stores cell identifiers and labels, whereas windowed kNN [19] retains feature vectors for neighbor searches. After an abrupt drift, the next \(W\) observations replace all pre-drift entries in the buffer. This turnover motivates setting \(\mathrm{max\_delay}=W\). For gradual drift, the transition width and the detection window are specified separately. In the observer comparison, Sliding Heatmap had lower processing times than windowed kNN on the tested streams, while kNN achieved higher predictive accuracy on all eight INSECTS variants.

\subsection{Detailed Description}
\label{subsection:sh_details}
Sliding Heatmap operates in a prequential setting on a stream of $(x_t, y_t)$. Before prediction, SH updates the cumulative minimum and maximum of each feature and uses them to map the input to a regular grid with \(G\) bins per dimension. When these bounds change, existing buffer entries retain their original cell identifiers. The bounds can therefore reflect observations that have already left the buffer.

The observer maintains class occurrence counts only for currently occupied grid cells using a sparse representation. In addition, a FIFO buffer stores, for each of the most recent $W$ observations, only its discretized cell identifier and class label. Full feature vectors are therefore not retained. When a new observation is processed, its cell count is updated and, once the window is full, the contribution associated with the oldest buffer entry is decremented. Empty cells are removed from the sparse representation, so the class counts contain only contributions from entries currently in the buffer.

Given an input $x_t$, the observer identifies its corresponding cell $c$. If $c$ contains observations, the prediction $\hat{y}_t$ is obtained as the majority class within the cell. If the cell is empty, an axis-aligned one-hop neighborhood is queried
by examining the immediately adjacent cell in both directions along each feature dimension, for at most $2d$ neighboring cells. If observations are found, the majority class across the neighborhood is returned. Otherwise, a fallback prediction is generated at random using a fixed seed, ensuring reproducibility.

After producing the prediction, the true label $y_t$ is observed, the counts of the corresponding cell are updated, and the sliding window is maintained by removing the oldest instance when necessary.
The computational cost of both prediction and update is constant with respect to the window size, as operations are performed on aggregated counts. Memory is bounded by the $W$ lightweight buffer entries and by the class counts associated with occupied grid cells; the full $G^d$ grid is never materialized. Unlike windowed kNN, SH does not retain full feature vectors for future neighbor searches.

\section{Experimental Evaluation}
\label{section: results}
\begin{figure*}[t]
    \centering
    \includegraphics[width=0.9\linewidth]{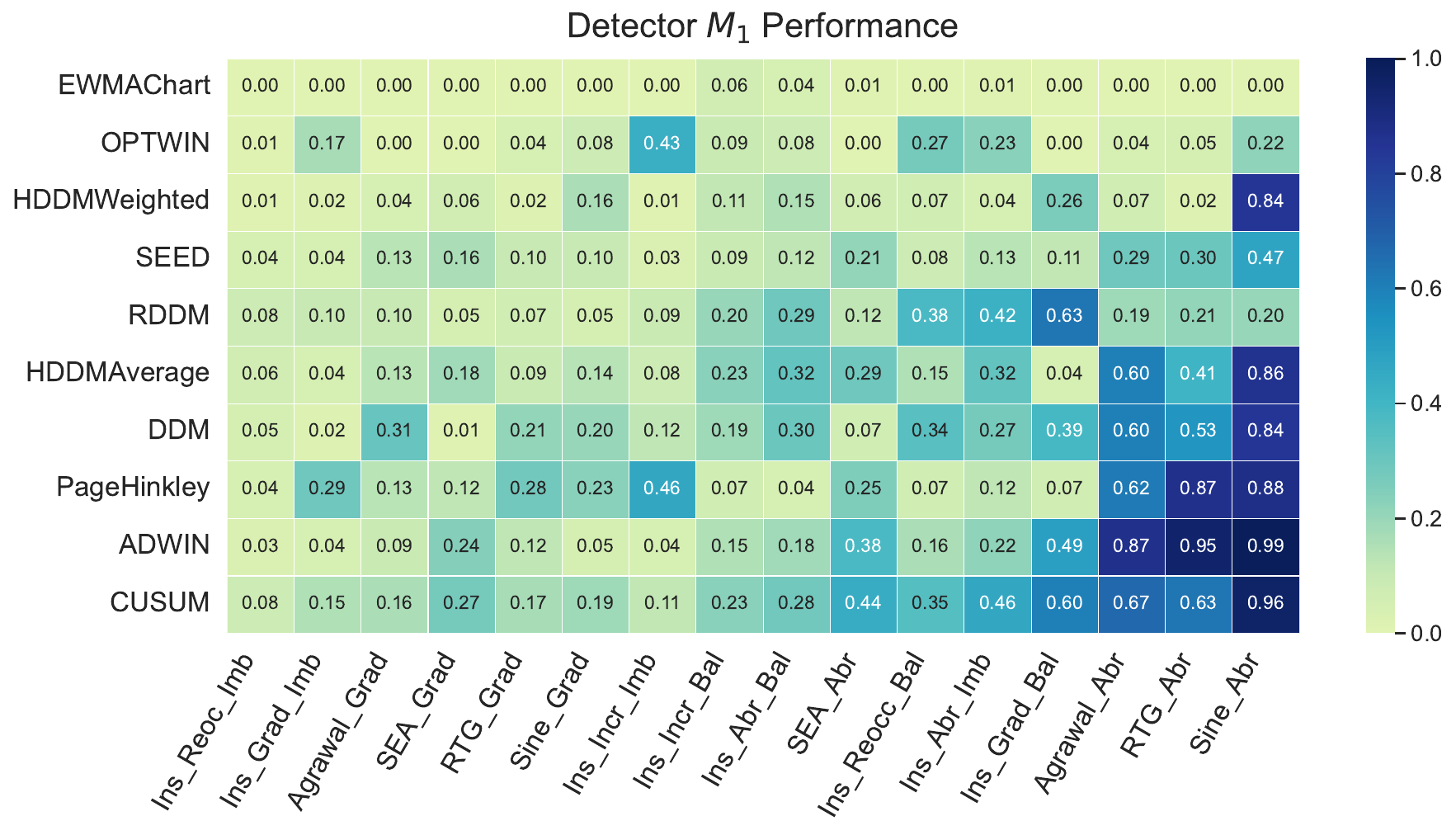}
    \caption{
        Heatmap of overall detector performance across all evaluated datasets, measured by the $M_1$ score. Rows and columns are ordered to visually enhance performance trends: detectors from worst (top) to best (bottom), and datasets from most challenging (left) to easiest (right). Color intensity reflects the magnitude of the $M_1$ score, with blue indicating higher performance.
    }
    \label{fig:heatmap}
\end{figure*}

\begin{figure}[b]
    \centering
    \includegraphics[width=0.5\textwidth]{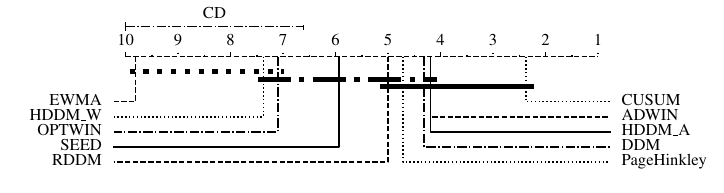}
    \caption{
        Critical difference diagram for $M_1$ scores, summarizing the performance across all datasets.
    }
    \label{fig:m1}
\end{figure}

This section presents the experimental evaluation of the proposed benchmark, including setup, datasets, and results. Experiments were conducted using CapyMOA~\cite{GomesLGSLHCB2025}, interfacing with MOA~\cite{BifetHKP2010}. All experiments rely on publicly available datasets and open-source software frameworks, including MOA and CapyMOA, whose original sources and licenses are acknowledged in the repository documentation. Sliding Heatmap was implemented in MOA and integrated into CapyMOA, ensuring consistent operation across all detectors. For each dataset, we evaluate 9 learner configurations obtained via grid search over window size and grid resolution, optimizing predictive accuracy while minimizing fallback predictions. This selection is independent of the detectors. Metrics are averaged across configurations, and \textit{max\_delay} is set equal to the window size.
We evaluate 10 detectors across 9 observer configurations and 16 datasets, resulting in over one thousand scenarios.

\subsection{Experimental Settings}
We benchmark ten concept drift detectors, covering diverse principles. ADWIN~\cite{BifetG2007}, which detects changes via adaptive windowing; CUSUM~\cite{Page1954}, based on cumulative deviations; DDM~\cite{GamaMCR2004}, which monitors error rate increases; EWMAChart~\cite{RossATH2012}, using exponentially weighted averages; HDDMAverage and HDDMWeighted~\cite{FriasCRMOC2014}, based on Hoeffding bounds; OPTWIN~\cite{TosiT2024}, which adapts window size dynamically; PageHinkley~\cite{Page1954}, for persistent change detection; RDDM~\cite{BarrosCGS2017}, extending DDM; and SEED~\cite{HuangKDP2014}, comparing short- and long-term statistics.

We evaluate detectors on real-world and synthetic datasets with annotated drift points. For real-world data, we consider INSECTS~\cite{SouzaRMB2020}, which provides diverse drift scenarios. Some streams exhibit challenging properties, such as long stationary periods or weak performance shifts, leading to potentially degenerate cases. Rather than filtering them, we retain all datasets to reflect real-world variability.
To complement this, we generate synthetic streams with controlled drift using: AgrawalGenerator~\cite{AgrawalIS1993} generates binary classification problems based on logical combinations of numeric and categorical attributes; RandomTreeGenerator~\cite{DomingosH2000} produces instances from a randomly generated decision tree, with controlled structure changes; SEA~\cite{StreetK2001} simple thresholds on numeric features; SineGenerator produces sinusoidal decision boundaries.
While our experiments focus on tabular data streams, the proposed framework is independent of the data modality. In high-dimensional settings such as images or videos, drift detection is typically applied on top of learned representations (e.g., embeddings), and our evaluation framework can be directly applied to the resulting performance signal.

\subsection{Experimental Results}
\begin{figure*}[t]
    \centering
    \begin{subfigure}{0.45\textwidth}
        \centering
        \includegraphics[width=\linewidth]{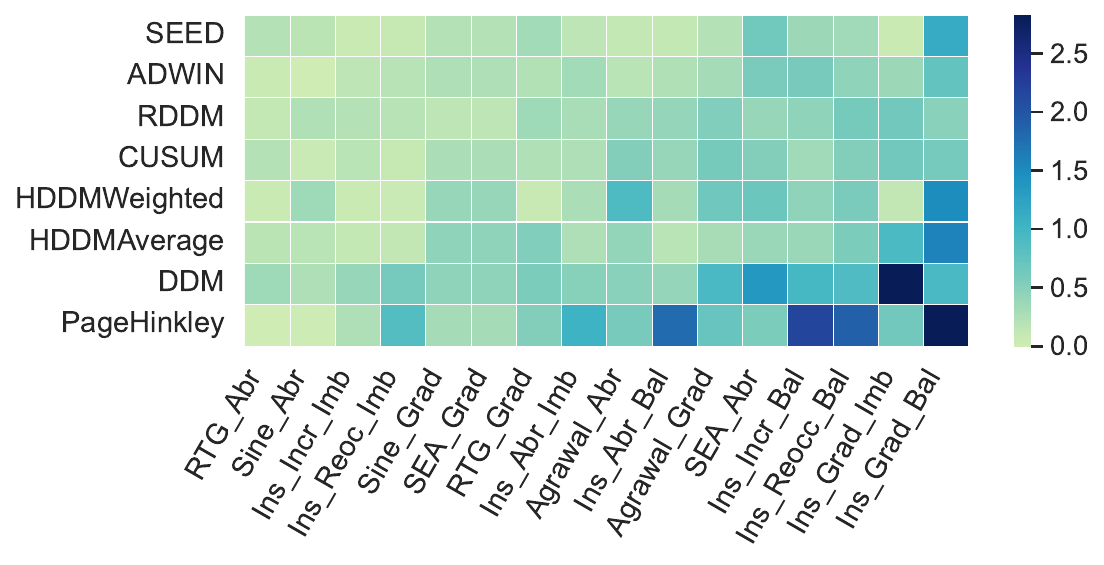}
        \caption{Coefficient of Variation (CV) of $M_1$}
        \label{fig:heatmap_CV}
    \end{subfigure}\hfill
    \begin{subfigure}{0.45\textwidth}
        \centering
        \includegraphics[width=\linewidth]{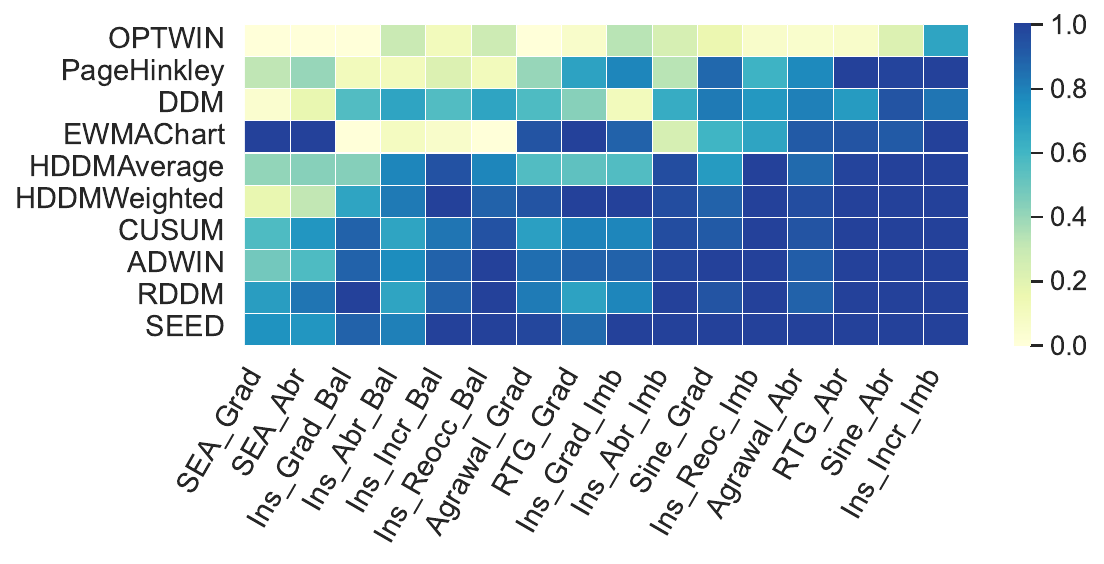}
        \caption{True Detection Rate (TDR)}
        \label{fig:heatmap_TDR}
    \end{subfigure}
    \begin{subfigure}{0.45\textwidth}
        \centering
        \includegraphics[width=\linewidth]{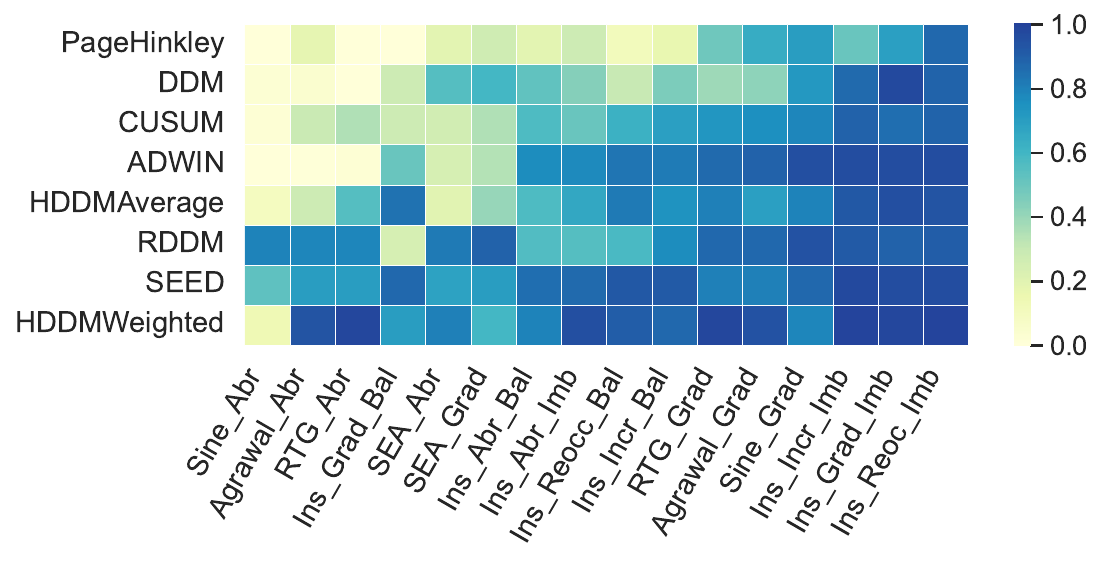}
        \caption{False Alarm Rate (FAR)}
        \label{fig:heatmap_FAR}
    \end{subfigure}\hfill
    \begin{subfigure}{0.45\textwidth}
        \centering
        \includegraphics[width=\linewidth]{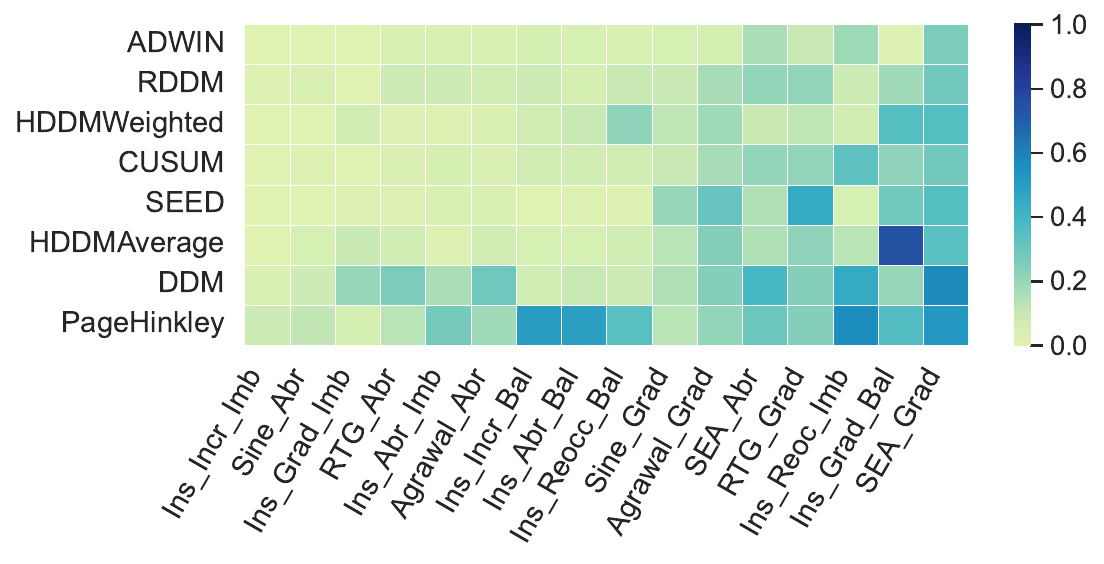}
        \caption{Mean Time to Detection normalized ($\mathrm{MTD}_{\mathrm{norm}}$)}
        \label{fig:heatmap_MTD}
    \end{subfigure}
    \caption{
    Comparison of drift detector performance across all evaluated datasets using complementary evaluation metrics.
    Rows and columns are ordered to highlight performance trends, with detectors and datasets ordered separately in each panel.
    For metrics where lower values indicate better performance (CV, FAR, and $\mathrm{MTD}_{\mathrm{norm}}$), higher-ranked detectors appear lower in the plot. For TDR, where higher values indicate better performance, higher-ranked detectors appear lower in the plot.
    }
    \label{fig:detailed_results}
\end{figure*}
Figure~\ref{fig:heatmap} provides a high-level overview of the overall performance of all concept drift detectors based on the composite metric $M_1$. Detectors are ordered by average score (bottom = better), and datasets by difficulty (left = harder). Dataset identifiers follow a compact notation: prefixes (e.g., Ins, SEA, RTG, Agrawal) denote data sources, while suffixes (\_Abr, \_Grad, \_Incr, \_Reocc, \_Bal, \_Imb) encode drift type and class balance. All detectors were evaluated using their default parameters reflecting a realistic use-case scenario where detectors are employed without extensive hyperparameter tuning. While some detectors may underperform due to non-optimal defaults, parameter adaptation was deliberately excluded to ensure comparability and avoid overfitting to specific datasets.

A clear structure emerges from the heatmap: datasets affected by gradual drift are predominantly on the left and those with abrupt drift on the right. This order was not enforced, but emerges naturally from the performance profiles. Detectors generally perform worse on gradual drift scenarios due to the absence of sudden changes in performance, and this reflects a known challenge in drift detection.

To complement the heatmap and support these findings with statistical evidence, Figure~\ref{fig:m1} shows the critical difference diagram for the $M_1$ scores. Detectors are ranked across all datasets, and those not connected by a horizontal bar are significantly different at $\alpha=0.05$. The diagram summarizes ranks across datasets. Horizontal bars connect detectors whose ranks are not significantly different under the Nemenyi test. These comparisons should be interpreted pairwise, since overlapping bars can connect subsets of detectors. As $M_1$ aggregates multiple dimensions, high scores do not guarantee uniformly strong performance. Trade-offs between detection, false alarms, and delay remain visible in the individual metrics. In the next section, we break down the statistical significance of detector performance along each individual metric.

Figure~\ref{fig:detailed_results} complements the aggregate analysis based on the composite score $M_1$ by decomposing performance into its constituent dimensions: detection accuracy, false alarms, detection delay, and stability. These detailed results confirm that detector performance cannot be adequately summarized by a single metric. Instead, effective evaluation requires balancing accuracy, timeliness, false alarms, and stability. The composite score $M_1$ provides a principled aggregation of these aspects, while the metric-specific heatmaps reported here offer complementary insight into the underlying trade-offs.

\subsection{Statistical Significance Analysis}
\begin{figure*}[t]
    \centering
    \subfloat[Coefficient of Variation (CV) of $M_1$\label{fig:CV}]{
        \includegraphics[width=0.45\textwidth]{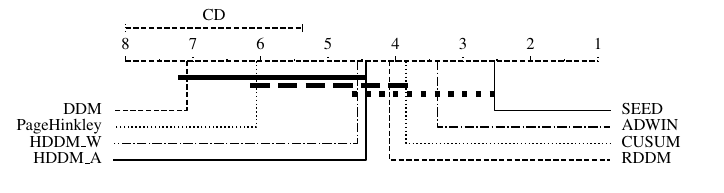}
    }
    \subfloat[True Detection Rate (TDR) \label{fig:TDR}]{
        \includegraphics[width=0.45\textwidth]{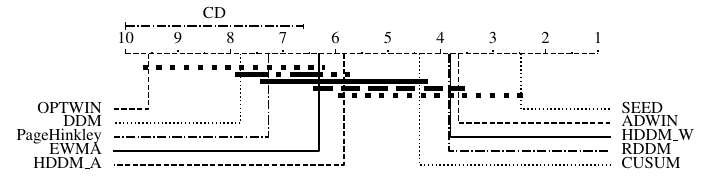}
    }
    \quad
    \subfloat[False Alarm Rate (FAR)\label{fig:FAR}]{
        \includegraphics[width=0.45\textwidth]{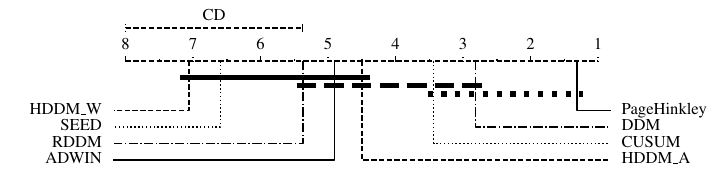}
    }
    \subfloat[MTD normalized ($\mathrm{MTD}_{\mathrm{norm}}$)\label{fig:MTD}]{
        \includegraphics[width=0.45\textwidth]{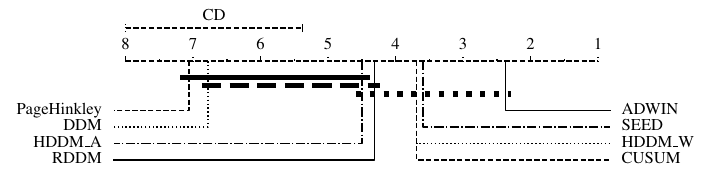}
    }
    \caption{Critical difference diagrams for the four evaluation metrics used. Horizontal bars indicate groups of detectors that are not significantly different according to the Nemenyi test ($\alpha = 0.05$).}
    \label{fig:CD}
\end{figure*}
To analyze detector behavior beyond $M_1$, we rank detectors across all datasets and summarize the results using critical difference (CD) diagrams. The \textbf{Friedman test} confirms statistically significant differences among detectors for all metrics; post-hoc pairwise comparisons are conducted using the Nemenyi test~\cite{Demsar2006} at $\alpha=0.05$. The resulting CD diagrams, shown in Figure~\ref{fig:CD}, reveal the statistical groupings of detectors. OPTWIN and EWMAChart are excluded from some analyses due to missing detections, which prevent reliable metric computation.

Figure~\ref{fig:CV} quantifies the sensitivity of each detector to different learner configurations. Lower values (right) correspond to more stable detectors. The resulting ranking shows moderate agreement, with detectors such as SEED appearing more robust, while DDM shows higher variability. Figure~\ref{fig:TDR} evaluates how frequently each detector correctly identifies true drift points, with fairly strong agreement. Detectors such as ADWIN achieve the highest rank, confirming the high sensitivity observed in the $M_1$ heatmap, while detection coverage varies across the detectors.

The precision of the detectors, assessed in Figure~\ref{fig:FAR}, yields solid agreement. Conservative detectors such as PageHinkley and DDM effectively limit false alarms, helping to explain their competitive $M_1$ scores, while SEED, RDDM, and ADWIN are more prone to false positives. Finally, Figure~\ref{fig:MTD} measures detection delay, with lower values being preferable. ADWIN, CUSUM, HDDM, and SEED emerge as the fastest detectors, consistent with their strong $M_1$ performance, whereas PageHinkley and DDM respond more slowly when evaluated only on detected drifts.

\section{Conclusions}
\label{section: conclusions}
We conducted a rigorous and comprehensive evaluation of commonly used concept drift detectors on a diverse suite of data streams. Beyond comparing existing methods, we propose a general evaluation framework that can support principled and reproducible comparisons. This framework is designed to be extensible and thus suitable for both current and future detectors. Our results show that rankings reported in the literature can change substantially under principled evaluation. Different detectors exhibit trade-offs between detection speed, precision, and alarm stability, and no method dominates across all metrics. Importantly, our results suggest that simply adding more datasets would not resolve the limitations of current benchmarks if evaluation protocols remain ill-defined. 

More broadly, our results suggest that concept drift detector evaluation remains an open methodological problem rather than a settled benchmarking task. We therefore argue that further research should first characterize detector behavior under controlled observational conditions that reduce learner-induced confounding. This does not imply that the learner is irrelevant; on the contrary, once detector behavior is better understood in isolation, learner dependence can be reintroduced explicitly and studied as a separate dimension of the evaluation problem. In this sense, the proposed framework should not be viewed as the final benchmarking protocol, but as an analysis tool for disentangling the factors that shape detector behavior.

Among the most relevant directions for \textbf{future work}, we identify the need to go beyond default configurations. In this study, we kept all detectors in their standard hyperparameter settings to ensure fairness and reproducibility. However, this likely penalized some methods and may have masked their true potential. Future studies could explore systematic hyperparameter tuning, adaptive configuration techniques, or even meta-learning approaches. Overall, this work lays the foundation for a standardized and reproducible evaluation of concept drift detectors, providing tools and baselines to guide the development of more robust and transparent solutions.


\begin{thebibliography}{00}
\bibitem{WidmerK1996} Widmer, Gerhard, and Miroslav Kubat. "Learning in the presence of concept drift and hidden contexts." Machine learning (1996): 69-101.

\bibitem{GamaZBPB2014} Gama, João, et al. "A survey on concept drift adaptation." ACM computing surveys (CSUR) 46.4 (2014): 1-37.

\bibitem{Tsymbal2004} Tsymbal, Alexey. "The problem of concept drift: definitions and related work." Computer Science Department, Trinity College Dublin (2004).

\bibitem{WebbHCNP2016} Webb, Geoffrey I., et al. "Characterizing concept drift." Data Mining and Knowledge Discovery 30.4 (2016): 964-994.

\bibitem{HoensPC2012} Hoens, T. Ryan, Robi Polikar, and Nitesh V. Chawla. "Learning from streaming data with concept drift and imbalance: an overview." Progress in artificial intelligence 1.1 (2012): 89-101.

\bibitem{KhamassiSHG2018} Khamassi, Imen, et al. "Discussion and review on evolving data streams and concept drift adapting." Evolving systems 9.1 (2018): 1-23.

\bibitem{LuLDGGZ2018} Lu, Jie, et al. "Learning under concept drift: A review." IEEE transactions on knowledge and data engineering 31.12 (2018): 2346-2363.

\bibitem{MahgoubMEA2022} Mahgoub, Mahmoud, et al. "Benchmarking concept drift detectors for online machine learning." International Conference on Model and Data Engineering. Cham: Springer Nature Switzerland, 2022.

\bibitem{HultenSD2001} Hulten, Geoff, Laurie Spencer, and Pedro Domingos. "Mining time-changing data streams." Proceedings of the seventh ACM SIGKDD international conference on Knowledge discovery and data mining. 2001.

\bibitem{BifetG2009} Bifet, Albert, and Ricard Gavalda. "Adaptive learning from evolving data streams." International symposium on intelligent data analysis. Berlin, Heidelberg: Springer Berlin Heidelberg, 2009.

\bibitem{Goncalves2014} Gonçalves Jr, Paulo M., et al. "A comparative study on concept drift detectors." Expert Systems with Applications 41.18 (2014): 8144-8156.

\bibitem{Bifet2017} Bifet, Albert. "Classifier concept drift detection and the illusion of progress." International conference on artificial intelligence and soft computing. Cham: Springer International Publishing, 2017.

\bibitem{HaugTK2022} Haug, Johannes, Effi Tramountani, and Gjergji Kasneci. "Standardized evaluation of machine learning methods for evolving data streams." arXiv preprint arXiv:2204.13625 (2022).

\bibitem{MahdiBS2024} Mahdi, Osama, Savitri Bevinakoppa, and Sarabjot Singh. "Diversity measure for concept drift detection in data streams." 2024 IEEE 9th International Conference on Engineering Technologies and Applied Sciences (ICETAS). IEEE, 2024.

\bibitem{TosiT2024} Tosi, Mauro DL, and Martin Theobald. "OPTWIN: Drift identification with optimal sub-windows." 2024 IEEE 40th International Conference on Data Engineering Workshops (ICDEW). IEEE, 2024.

\bibitem{SouzaRMB2020} Souza, Vinicius MA, et al. "Challenges in benchmarking stream learning algorithms with real-world data." Data Mining and Knowledge Discovery 34.6 (2020): 1805-1858.

\bibitem{Zliobaite2010} Žliobaitė, Indrė. "Learning under concept drift: an overview." arXiv preprint arXiv:1010.4784 (2010).

\bibitem{DevroyeGL2013} Devroye, Luc, László Györfi, and Gábor Lugosi. A probabilistic theory of pattern recognition. Vol. 31. New York: springer, 1996.

\bibitem{ReadBPH2012} Read, Jesse, et al. "Batch-incremental versus instance-incremental learning in dynamic and evolving data." International symposium on intelligent data analysis. Berlin, Heidelberg: Springer Berlin Heidelberg, 2012.

\bibitem{GomesLGSLHCB2025} Gomes, Heitor Murilo, et al. "Capymoa: Efficient machine learning for data streams in python." (2025).

\bibitem{BifetHKP2010} Bifet, Albert, et al. "Moa: Massive online analysis, a framework for stream classification and clustering." Proceedings of the first workshop on applications of pattern analysis. PMLR, 2010.

\bibitem{BifetG2007} Bifet, Albert, and Ricard Gavalda. "Learning from time-changing data with adaptive windowing." Proceedings of the 2007 SIAM international conference on data mining, 2007.

\bibitem{Page1954} Page, Ewan S. "Continuous inspection schemes." Biometrika (1954).

\bibitem{GamaMCR2004} Gama, Joao, et al. "Learning with drift detection." Brazilian symposium on artificial intelligence. Springer Berlin Heidelberg, 2004.

\bibitem{RossATH2012} Ross, Gordon J., et al. "Exponentially weighted moving average charts for detecting concept drift." Pattern recognition letters (2012): 191-198.

\bibitem{FriasCRMOC2014} Frias-Blanco, Isvani, et al. "Online and non-parametric drift detection methods based on Hoeffding’s bounds." IEEE Transactions on Knowledge and Data Engineering 27.3 (2014): 810-823.

\bibitem{BarrosCGS2017} Barros, Roberto SM, et al. "RDDM: Reactive drift detection method." Expert Systems with Applications 90 (2017): 344-355.

\bibitem{HuangKDP2014} Huang, David Tse Jung, et al. "Detecting volatility shift in data streams." 2014 IEEE International Conference on Data Mining. IEEE, 2014.

\bibitem{AgrawalIS1993} Agrawal, Rakesh, Tomasz Imielinski, and Arun Swami. "Database mining: A performance perspective." IEEE transactions on knowledge and data engineering 5.6 (1993): 914-925.

\bibitem{DomingosH2000} Domingos, Pedro, and Geoff Hulten. "Mining high-speed data streams." Proceedings of the sixth ACM SIGKDD international conference on Knowledge discovery and data mining. 2000.

\bibitem{StreetK2001} Street, W. Nick, and YongSeog Kim. "A streaming ensemble algorithm (SEA) for large-scale classification." Proceedings of the seventh ACM SIGKDD, 2001.

\bibitem{Demsar2006} Demšar, Janez. "Statistical comparisons of classifiers over multiple data sets." Journal of Machine learning research 7.Jan (2006): 1-30.

\end{thebibliography}
\end{document}